\documentclass{article}
\usepackage{spconf,amsmath,graphicx,hyperref}
\ninept

\usepackage{color}
\usepackage{epsfig}
\usepackage{graphicx}

\usepackage{booktabs}  %
\usepackage{tabularx}               %
\usepackage{ltablex}

\newcolumntype{C}{>{\centering\arraybackslash}X}
\usepackage{multirow}               %
\usepackage{diagbox}                %
\usepackage{hhline}                 %

\usepackage{extarrows}
\usepackage{makecell}
\usepackage{colortbl}
\usepackage{longtable}
\usepackage[most]{tcolorbox}
\tcbuselibrary{skins} %

\usepackage{amssymb}
\usepackage{amsthm}
\usepackage{amsfonts}

\usepackage{adjustbox}
\usepackage{array}
\usepackage{floatflt}

\usepackage{bm}
\usepackage{nicefrac}
\usepackage{microtype}

\usepackage{changepage}
\usepackage{extramarks}
\usepackage{fancyhdr}
\usepackage{lastpage}
\usepackage{soul}
\usepackage{xspace}

\usepackage{arydshln} %

\usepackage{enumerate}
\usepackage{enumitem}  %

\usepackage{pifont} %

\usepackage{algpseudocode}
\usepackage[ruled,vlined]{algorithm2e} %

\usepackage[symbol]{footmisc}

\usepackage[caption=false]{subfig}

\usepackage{scalefnt}

\usepackage{fontawesome5}

\usepackage{siunitx}

\usepackage{listings}

\newcolumntype{L}[1]{>{\raggedright\let\newline\\\arraybackslash\hspace{0pt}}m{#1}}
\newcolumntype{R}[1]{>{\raggedleft\let\newline\\\arraybackslash\hspace{0pt}}m{#1}}

\newcommand{\ignore}[1]{}

\makeatletter
\DeclareRobustCommand\onedot{\futurelet\@let@token\@onedot}
\def\@onedot{\ifx\@let@token.\else.\null\fi\xspace}

\makeatother

\definecolor{Emerald}{HTML}{50C878}

\definecolor{bestBase}{HTML}{D44478}
\definecolor{secondBase}{HTML}{FFAA33}
\definecolor{thirdBase}{HTML}{FFDD33}

\colorlet{bestLight}{bestBase!20!white}
\colorlet{secondLight}{secondBase!20!white}
\colorlet{thirdLight}{thirdBase!20!white}

\definecolor{MyBlue}{rgb}{0.46, 0.50, 0.61}
\definecolor{MyDarkBlue}{rgb}{0,0.08,0.8}
\definecolor{MyDarkGreen}{RGB}{45,155,45}
\definecolor{MyDarkRed}{rgb}{0.8,0.02,0.02}
\definecolor{MyOrange}{rgb}{1.0, 0.4, 0.2}
\definecolor{MyPurple}{RGB}{111,0,255}
\definecolor{MyRed}{rgb}{0.8,0.0,0.0}
\definecolor{MyGold}{rgb}{0.75,0.6,0.12}
\definecolor{MyDarkgray}{rgb}{0.66, 0.66, 0.66}
\definecolor{MyBrown}{rgb}{0.65, 0.16, 0.16}
\definecolor{MyMutedRose}{rgb}{0.58, 0.29, 0.35}
\definecolor{JiayuanColor}{rgb}{0.60,0.43,0.48}
\definecolor{erranColor}{rgb}{24, 40, 113}

\definecolor{citecolor}{HTML}{696FAD}

\DeclareRobustCommand{\modelname}{\textls[-15]{TIAO}\xspace}

\newif\ifpropositionfirstitem
\propositionfirstitemtrue

\definecolor{bggray}{HTML}{F5F5F5}
\definecolor{pvdblue}{HTML}{DAE8FC}
\definecolor{RoseQuartzBg}{HTML}{F7CAC9}
\definecolor{RoseQuartz}{HTML}{F5A798}
\definecolor{Serenity}{HTML}{92A8D1}
\definecolor{OrangeRed}{rgb}{1.0, 0.27, 0.0}
\definecolor{RoyalBlue}{cmyk}{1, 0.50, 0, 0}
\definecolor{Turquoise}{HTML}{0F4C81}
\definecolor{mint}{rgb}{0.24, 0.71, 0.54}

\newdimen\abovecrulesep
\newdimen\belowcrulesep
\makeatletter
\patchcmd{\@@@cmidrule}{\aboverulesep}{\abovecrulesep}{}{}
\patchcmd{\@xcmidrule}{\belowrulesep}{\belowcrulesep}{}{}
\makeatother

\definecolor{mybluetitle}{HTML}{4B527E} %

\definecolor{mygreen}{RGB}{0,150,0}
\definecolor{boxbackground}{HTML}{F0F7FF}  %
\definecolor{boxborder}{HTML}{D0D9E5}      %
\definecolor{accentblue}{HTML}{4A86E8}     %
\definecolor{lightblue}{HTML}{EEF3FF}  %
\definecolor{bordergray}{HTML}{CCCCCC}  %
\definecolor{headerblue}{HTML}{2C5AA0}  %

\definecolor{lavenderframe}{HTML}{E6E6FA}  %
\definecolor{lighterlav}{HTML}{F5F5FF}  %
\definecolor{codegray}{rgb}{0.5,0.5,0.5}  %
\definecolor{codepurple}{HTML}{483D8B}  %
\definecolor{backcolour}{HTML}{F5F5FF}  %
\lstdefinestyle{mystyle}{
    backgroundcolor=\color{backcolour},
    commentstyle=\color{headerblue},
    keywordstyle=\color{codepurple},
    numberstyle=\tiny\color{codegray},
    stringstyle=\color{codepurple},
    basicstyle=\ttfamily\scriptsize,
    breakatwhitespace=false,
    breaklines=true,
    captionpos=b,
    keepspaces=true,
    frame=none,
    numbersep=5pt,
    showspaces=false,
    showstringspaces=false,
    showtabs=false,
    tabsize=2
}

\definecolor{jsonkey}{RGB}{44, 130, 201}     %
\definecolor{jsonstring}{RGB}{255, 140, 0}   %
\definecolor{jsonnumber}{RGB}{34, 139, 34}   %

\lstdefinelanguage{json}{
    basicstyle=\ttfamily\small,
    numbers=left,
    numberstyle=\tiny\color{gray},
    stepnumber=1,
    numbersep=5pt,
    showstringspaces=false,
    breaklines=true,
    frame=none,
    backgroundcolor=\color{gray!5},
    literate=
     *{:}{{{\color{jsonkey}:}}}{1}
      {,}{{{\color{jsonkey},}}}{1}
      {"}{{{\color{jsonstring}"}}}{1}
      {[}{{{\color{jsonkey}[}}}{1}
      {]}{{{\color{jsonkey}]}}}{1}
      {0}{{{\color{jsonnumber}0}}}{1}
      {1}{{{\color{jsonnumber}1}}}{1}
      {2}{{{\color{jsonnumber}2}}}{1}
      {3}{{{\color{jsonnumber}3}}}{1}
      {4}{{{\color{jsonnumber}4}}}{1}
      {5}{{{\color{jsonnumber}5}}}{1}
      {6}{{{\color{jsonnumber}6}}}{1}
      {7}{{{\color{jsonnumber}7}}}{1}
      {8}{{{\color{jsonnumber}8}}}{1}
      {9}{{{\color{jsonnumber}9}}}{1}
}

\newtcblisting{jsonbox}{
  listing engine=listings,
  colback=gray!3!white,
  colframe=gray!75!black,
  boxrule=0.4mm,
  arc=2mm,
  outer arc=2mm,
  breakable,
  enhanced,
  listing only,
  listing options={language=json}
}

\newtcolorbox{promptbox}[2][]{ %
    enhanced,
    breakable,
    boxsep=5pt,
    left=9pt,
    right=7pt,
    top=5pt,
    bottom=5pt,
    colback=boxbackground,
    colframe=boxborder,
    boxrule=0.5pt,
    arc=4pt,
    frame hidden, %
    borderline west={3pt}{0pt}{accentblue},
    shadow={0.5pt}{0.5pt}{1.5pt}{black!10},
    fontupper=\normalsize,
    title=#2, %
    colbacktitle=accentblue, %
    coltitle=white,         %
    fonttitle={\fontsize{9}{11}\selectfont\bfseries}, %
    attach boxed title to top left={yshift=-2.5mm, xshift=3.2mm},
    boxed title style={
        enhanced,
        left=3pt,
        right=3pt,
        top=1pt,    %
        bottom=1pt, %
        boxsep=2pt,
        arc=3pt,
        boxrule=0pt,
        colback=accentblue,
    },
    #1 %
}

\newtcolorbox{notitlepromptbox}[1][]{
    enhanced,
    breakable,
    boxsep=5pt,          %
    left=9pt,            %
    right=7pt,           %
    top=5pt,             %
    bottom=5pt,          %
    colback=boxbackground,
    colframe=boxborder,
    boxrule=0.5pt,
    arc=4pt,             %
    frame hidden,
    borderline west={3pt}{0pt}{accentblue},  %
    shadow={0.5pt}{0.5pt}{1.5pt}{black!10},  %
    notitle,
    fontupper=\normalsize,    %
    #1
}

\newtcolorbox{onebox}[2][]{
    enhanced, 
    center title,
    left*=0pt, right*=0pt,
    boxsep=2pt, left=5pt, right=5pt,
    skin first=enhanced,
    skin middle=enhanced,
    skin last=enhanced,
    colframe = mybluetitle!90,
  colback  = mybluetitle!10,
    fonttitle=\bfseries\rmfamily\fontfamily{phv}\selectfont,
    title={\footnotesize\strut{#2}  \refstepcounter{subsubsection} \addcontentsline{toc}{subsubsection}{\string\numberline{\thesubsubsection}#2}
    },
    #1
    }

\colorlet{osfirst}{teal!50}
\colorlet{ossecond}{teal!30}
\colorlet{osthird}{teal!10}
\colorlet{lavenderfirst}{violet!50}
\colorlet{lavendersecond}{violet!30}
\colorlet{lavenderthird}{violet!10}

\theoremstyle{plain}

\theoremstyle{definition}

\theoremstyle{remark}

\definecolor{lightgray}{rgb}{0.88, 0.92, 0.98}
\definecolor{defblue}{rgb}{0.1843, 0.3333, 0.6}
\definecolor{defred}{rgb}{0.88, 0.2510, 0.3294}

\definecolor{green1}{rgb}{ 0.910,  0.953,  0.855}
\definecolor{green2}{rgb}{0.82,  0.902,  0.710}
\definecolor{green3}{rgb}{0.713,  0.903,  0.648}
\definecolor{green4}{rgb}{ 0.725,  0.855,  0.561}

\definecolor{defyellow}{rgb}{1,  0.983,  0.717}
\definecolor{defyellowtext}{rgb}{1,  0.851,  0.438}

\title{\modelname: Token Importance-Aware Policy Optimization for Text Summarization}
\name{Qixiu Li$^{1,\dagger}$
        \! Chenlong Bao$^{1,\dagger}$
        \! Xiang Zhu$^{1,*}$
        \! Xiaoyong Li$^{1,*}$\thanks{$\dagger$ Equal Contribution. *Corresponding Author.}
        \! Ruixin Cao$^{1}$
        \! Shukai Chen$^{1}$
        \! Zhenxiong Zhou$^{1}$
        }

\address{$^{1}$National University of Defense Technology}

\begin{document}
%
\maketitle

\begin{abstract}
Text summarization requires models to condense content while preserving key qualities such as consistency and coherence. Large language models (LLMs) have shown strong performance on this task and can be further improved through reinforcement learning (RL). However, most existing methods apply reward signals directly to undifferentiated token sequences, overlooking the varying importance of individual tokens to word and sentence level quality in summarization. In this paper, we propose Token Importance-Aware Policy Optimization (\modelname), a novel reinforcement learning strategy that explicitly leverages token-importance awareness. Specifically, \modelname identifies core tokens based on token dependency and reweights a trajectory's advantage according to its overall dependencies. Experiments on the real world dataset show that our \modelname achieves highly competitive results, and that a 7B foundation model enhanced by \modelname performs comparably to GPT-4 and GPT-5-nano. Code is available at~\href{https://github.com/TechCloud-x/TIAO}{https://github.com/TechCloud-x/TIAO}.
\end{abstract}
\begin{keywords}
Reinforcement Learning, Text Summarization
\end{keywords}

\vspace{-1.2em}
\section{Introduction}\label{sec:intro}
\vspace{-0.9em}
Text summarization is a selective information compression problem: a system must shorten a source document while preserving salient entities, numbers, events, and relations, and while maintaining coherence, relevance, fluency, and factual consistency. Pre-trained sequence-to-sequence models and recent large language models (LLMs) have substantially improved abstractive summarization quality~\cite{fan2026eva,liang2025quantifying,zhang2024benchmarking}. Yet factuality studies show that fluent summaries may still hallucinate or distort source-supported content~\cite{wang2026raptm,maynez2020faithfulness,pagnoni2021understanding}; once key evidence is dropped during compression, fluent surface realization cannot recover it. This has motivated multi-dimensional evaluators for consistency, coherence, relevance, and fluency~\cite{zhong2022towards,laban2022summac,liu2023geval}. Reinforcement learning (RL) offers a direct way to optimize such non-differentiable quality signals, from human-feedback summarization~\cite{stiennon2020learning} to recent multi-objective reward balancing policy HVO~\cite{song2026balancing}. These methods mainly improve what reward should define a better summary.

However, summarization policy optimization also faces a credit-resolution mismatch. Autoregressive generation is a token-level sequence decision process, where the policy selects each token conditioned on the source and the generated prefix. In contrast, summary quality is usually evaluated only after the whole sequence is completed. In a GRPO-style objective~\cite{shao2024deepseekmath}, the same trajectory advantage is commonly multiplied by token-wise policy ratios across positions. This distinguishes better and worse summary trajectories, but cannot identify which positions actually depend on the source document or determine factual coverage. The resulting reward broadcast can dilute useful gradients over many low-information tokens.

Recent fine-grained RL methods partially address this limitation. Process reward models and step-wise verifiers provide denser feedback for reasoning traces~\cite{lightman2024verify,wang2024mathshepherd}, while reward redistribution decomposes holistic feedback into token-level rewards~\cite{li2025red}. Entropy-based analysis further shows that a small fraction of uncertain tokens can dominate RL gains in reasoning models~\cite{wang2026beyond}. Optimization-side advances such as DAPO~\cite{yu2026dapo} and SAPO~\cite{gao2025soft} improve large-scale LLM RL through token-level design, dynamic sampling, and token-adaptive update control. Nevertheless, these approaches often rely on auxiliary reward models, process annotations, or output-side proxy signals, and are largely developed for mathematical or general reasoning. For summarization, the central question is different: whether an output token is genuinely supported by source evidence.

\begin{figure}[t!]
  \begin{center}
\includegraphics[width=.86\linewidth]{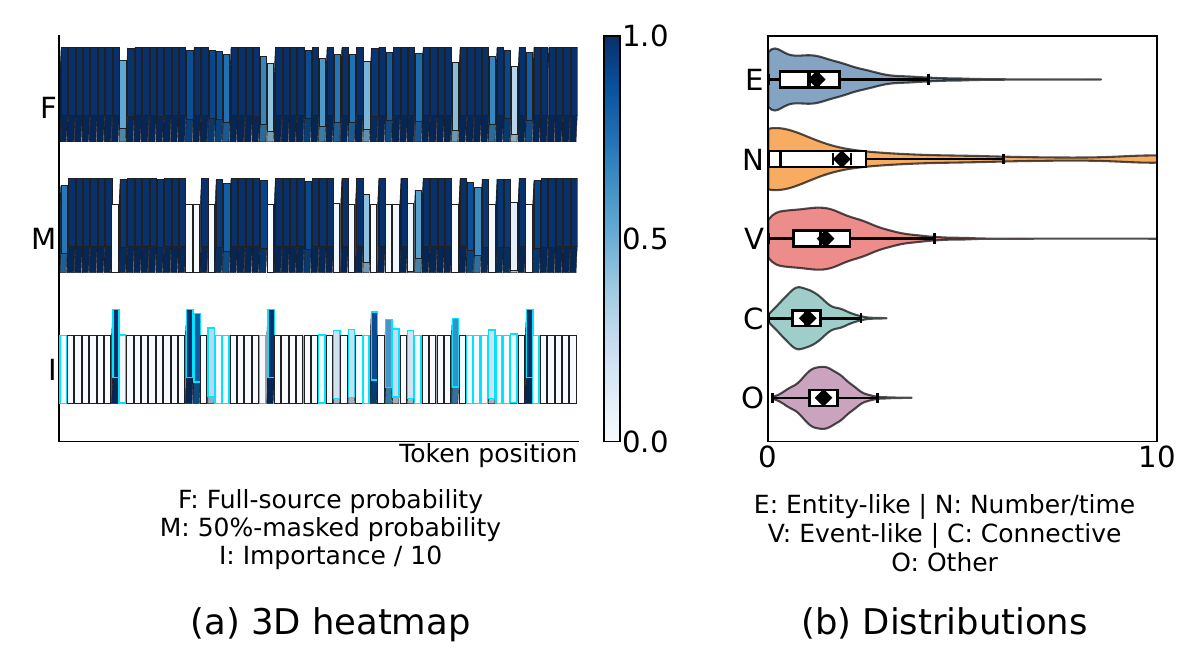}
  \end{center}
    \vspace{-22pt}
  \caption{\modelname token dependency analysis under source masking. (a) compares full-source and masked-source token probabilities and highlights source-dependent update positions. (b) summarizes trajectory-level dependency patterns across token categories.}
  \vspace{-10pt}
  \label{fig:teaser}
  \vspace{-11pt}
\end{figure}

This paper targets this missing link. Under multi-dimensional summarization rewards, existing algorithms still lack source-dependency-aware credit assignment that can align sequence-level quality feedback with token-level evidence usage. Such a mechanism should distinguish source-grounded trajectories from language-prior shortcuts and, within the same trajectory, assign stronger learning pressure to tokens whose probabilities are sensitive to the source document. Importantly (Fig.~\ref{fig:teaser}), this should be achieved without training additional token critic. Better rewards specify what to learn; our goal is to decide where and how strongly the policy should learn.

To this end, we propose \textbf{\underline{T}}oken \textbf{\underline{I}}mportance-\textbf{\underline{A}}ware Policy \textbf{\underline{O}}ptimization (\textbf{\modelname}), a source-sensitive policy optimization method for text summarization. \modelname forms a dual-scale optimization loop: source perturbation estimates token dependency; token dependencies are aggregated into trajectory-level source dependency; trajectory advantages are scaled by this dependency; and token updates are focused on source-sensitive positions. Our contributions are threefold. \ding{182} We define a source-dependency measure for summary output tokens based on probability changes under counterfactual source masking. \ding{183} We introduce a trajectory-token credit assignment strategy that jointly reweights trajectory advantages and concentrates token-level policy updates. \ding{184} We validate \modelname on CNN/DailyMail~\cite{nallapati2016abstractive}, demonstrating improvements in summary quality, training stability.

\begin{figure*}[ht]
\vspace{-10pt}
  \begin{center}
  \includegraphics[width=0.9\linewidth]{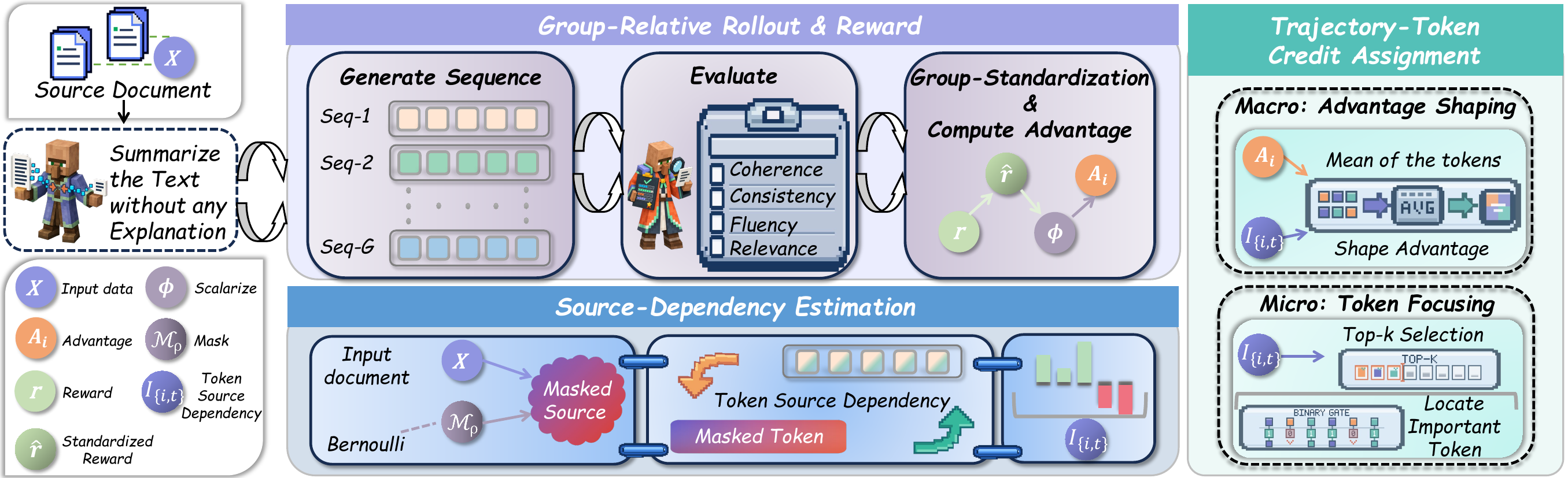}
  \end{center}
  \vspace{-20pt}
  \caption{\small The logic implementation of our proposed \modelname.}
  \vspace{-15pt}
  \label{fig:main}
\end{figure*}
\section{Methodology}\label{sec:method}

\subsection{Overview}
\noindent We propose Token Importance-Aware Policy Optimization (\textbf{\modelname}), a source-sensitive reinforcement learning framework for abstractive summarization. As shown in Fig.~\ref{fig:main}, \modelname keeps the standard group-relative rollout and multi-dimensional summarization rewards, but changes how the resulting learning signal is assigned. For each sampled summary, we perturb the source document and re-score the same output tokens under the original and perturbed sources. The induced probability shift estimates how strongly each output token depends on source evidence. \modelname then uses this signal at two levels: it reshapes the trajectory advantage according to the summary's overall source dependency, and it filters token-level policy gradients toward the most source-sensitive output positions. Thus, the reward still defines what summary is better, while \modelname determines where the policy should learn from it.

\subsection{Problem Formulation}
\noindent Given a document $x$ and a reference summary $y^{\star}$, a policy $\pi_{\theta}$ generates an abstractive summary $y=(y_1,\ldots,y_T)$ autoregressively:
\begin{equation}
    y_t \sim \pi_{\theta}(\cdot \mid x,y_{<t}), \quad t=1,\ldots,T .
\end{equation}
Text summarization policy optimization is therefore a sequence decision problem with token-level actions and sequence-level feedback. Since generated summaries have variable lengths, we define a valid-token mask $c_{i,t}\in\{0,1\}$ for the $i$-th sampled trajectory, where tokens after the first end-of-sequence marker are excluded and $T_i=\sum_t c_{i,t}$. Let $\mathbf{r}(x,y)\in\mathbb{R}^{D}$ denote a multi-dimensional evaluator, where the dimensions correspond to coherence, consistency, fluency, and relevance in our implementation. A scalar trajectory reward is obtained through an external aggregation function $\Phi$, calculated as:
\begin{equation}
    R(x,y)=\Phi(\mathbf{r}(x,y)).
\end{equation}
\modelname is independent of the specific evaluator and aggregation rule. Its role is to convert the resulting trajectory-level signal into source-aware trajectory and token credits without an additional token critic.

\subsection{Multi-Dimensional Group-Relative Policy Optimization}
\noindent For each document $x$, the old policy $\pi_{\theta_{\mathrm{old}}}$ samples a group of $G$ summaries $\{y_i\}_{i=1}^{G}$. Each summary first receives a reward vector $\mathbf{r}_i=[r_{i,1},\ldots,r_{i,D}]$. To make different dimensions comparable before scalarization, we write a group-wise standardized score as:
\begin{equation}
    \hat{r}_{i,d}=
    \frac{r_{i,d}-\mu_d}{\sigma_d+\epsilon_d},\quad
    \mu_d=\frac{1}{G}\sum_{j=1}^{G}r_{j,d}.
\end{equation}
The scalar reward is then obtained as $R_i=\Phi(\hat{\mathbf{r}}_i;\bm{\lambda})$, where $\bm{\lambda}\in\Delta^{D-1}$ denotes non-negative dimension preferences and $\Phi$ may represent a linear or non-linear aggregation. \modelname treats this reward construction as an external evaluator and modifies only the credit assigned to generated tokens. Following the group-relative policy optimization paradigm, we normalize scalar rewards within the group, which can be mathematically formulated as:
\begin{equation}
    A_i=\frac{R_i-\mu_R}{\sigma_R+\epsilon},\quad
    \mu_R=\frac{1}{G}\sum_{j=1}^{G}R_j ,
\end{equation}
where $\sigma_R$ is the within-group reward standard deviation and $\epsilon$ is a small constant. The token-wise policy ratio is calculated as:
\begin{equation}
    w_{i,t}(\theta)=
    \frac{\pi_{\theta}(y_{i,t}\mid x,y_{i,<t})}
    {\pi_{\theta_{\mathrm{old}}}(y_{i,t}\mid x,y_{i,<t})}.
\end{equation}
The corresponding unclipped token contribution can be written as $g_{i,t}=c_{i,t}w_{i,t}(\theta)A_i$, which shows the credit-resolution issue explicitly: all valid positions inherit the same trajectory-level scalar. In conventional GRPO-style training, this is efficient, but it treats factual content tokens, discourse markers, and low-information function words as equally responsible for the sequence-level reward.

\subsection{Source-Dependency Estimation for Output Tokens}
\noindent \modelname estimates token importance by measuring whether the probability of an already generated token changes when source evidence is partially removed. Let the source be $x=(x_1,\ldots,x_N)$. For each sampled trajectory, we draw an independent source mask as:
\begin{equation}
    b_{i,n}\sim\operatorname{Bernoulli}(1-\rho),\quad
    \tilde{x}_{i,n}=
    \begin{cases}
        x_n, & b_{i,n}=1,\\
        \mathtt{[MASK]}, & b_{i,n}=0,
    \end{cases}
\end{equation}
where $\rho\in(0,1)$ controls the perturbation strength. This defines the counterfactual source $\tilde{x}_i=\mathcal{M}_{\rho}(x;\mathbf{b}_i)$ while keeping the generated prefix $y_{i,<t}$ unchanged. We then compute teacher-forced log probabilities under the rollout policy. The  process is as follow:
\begin{equation}
    \ell^{F}_{i,t}=\log\pi_{\theta_{\mathrm{old}}}(y_{i,t}\mid x,y_{i,<t}),
    \quad
    \ell^{M}_{i,t}=\log\pi_{\theta_{\mathrm{old}}}(y_{i,t}\mid \tilde{x}_i,y_{i,<t}).
\end{equation}
We define the token source dependency as a sampled-token low-variance estimate of the conditional divergence between the full-source and masked-source predictions, which can be expressed as:
\begin{equation}
    d_{i,t}=\operatorname{clip}(\ell^{M}_{i,t}-\ell^{F}_{i,t},-\tau,\tau),
    \quad
    I_{i,t}=c_{i,t}\left[\exp(d_{i,t})-d_{i,t}-1\right].
\end{equation}
A larger $I_{i,t}$ indicates that $y_{i,t}$ is more sensitive to the source document. The clipping threshold $\tau$ only stabilizes extreme log-probability shifts and does not introduce any trainable critic. This formulation focuses on source support rather than output-side uncertainty alone: a token is important only when removing source evidence changes the model's confidence in producing it.

\subsection{Trajectory--Token Credit Assignment}
\noindent Based on the dependency scores, \modelname reshapes the learning signal at both macro and micro levels.

\noindent\textbf{Macro-level: trajectory advantage shaping.}
For each generated summary, we aggregate token dependencies into a trajectory dependency score. This operation can be mathematically formulated as:
\begin{equation}
    S_i=\frac{\sum_t c_{i,t}I_{i,t}}{\sum_t c_{i,t}+\delta_T},
\end{equation}
where $\delta_T$ prevents division by zero for degenerate completions. We then compute a mean-preserving positive scale over the rollout batch $\mathcal{B}$, this operation can be formulated as:
\begin{equation}
    \alpha_i=\frac{S_i}{\bar{S}+\delta},\quad
    \bar{S}=\frac{1}{|\mathcal{B}|}\sum_{j\in\mathcal{B}}S_j ,
\end{equation}
where $\delta$ prevents numerical instability. The shaped advantage is:
\begin{equation}
    \tilde{A}_i=\alpha_i A_i .
\end{equation}
This scaling amplifies updates for high-reward summaries that are strongly grounded in the source, while also applying stronger corrective pressure to low-reward summaries whose errors occur in source-dependent regions.

\noindent\textbf{Micro-level: token update focusing.}
Within each trajectory, we select the top-$\kappa$ proportion of valid output tokens according to $I_{i,t}$, where $\kappa\in(0,1)$. Let $\operatorname{rank}_{\downarrow}(I_{i,t})$ be the descending rank of token $t$ among valid tokens in the same trajectory, with invalid positions assigned infinite rank. The selected index set and binary token gate are computed mathematically as:
\begin{equation}
    \mathcal{K}_i=
    \{t\mid c_{i,t}=1,\operatorname{rank}_{\downarrow}(I_{i,t})
    \leq \lceil \kappa T_i\rceil\},\quad
    m_{i,t}=c_{i,t}\mathbb{I}(t\in\mathcal{K}_i).
\end{equation}
The gate is detached from policy optimization and serves only as a credit-assignment mask. By concentrating the policy-gradient term on source-sensitive tokens, \modelname reduces the diffusion of sequence-level rewards over generic or weakly grounded positions.

\subsection{Training Objective}
\noindent Integrating trajectory advantage shaping and token update focusing gives the final \modelname objective, which can be formulated as:
\begin{equation}
\begin{aligned}
\mathcal{L}_{\mathrm{TIAO}}(\theta)
&= \mathbb{E}_{x\sim\mathcal{D},\,
   \mathbf{y}\sim\pi_{\theta_{\mathrm{old}}}^{G}(\cdot\mid x)}
   \Bigg[
   \frac{1}{G}\sum_{i=1}^{G}
   \frac{1}{T_i}\sum_{t} \\
&\; m_{i,t}\min\big(
   w_{i,t}(\theta)\tilde{A}_i,
   \operatorname{clip}(w_{i,t}(\theta),1-\varepsilon,1+\varepsilon)\tilde{A}_i
   \big)
   \Bigg].
\end{aligned}
\end{equation}
Equivalently, \modelname optimizes an active surrogate as follow:
\begin{equation}
    \psi_{i,t}(\theta)=
    m_{i,t}\min\left(
    w_{i,t}(\theta)\tilde{A}_i,
    \operatorname{clip}(w_{i,t}(\theta),1-\varepsilon,1+\varepsilon)\tilde{A}_i
    \right),
\end{equation}
whose support is the source-sensitive set $\mathcal{A}_i=\{t\mid m_{i,t}=1\}$. Thus, the gradient estimator excludes inactive positions before summation while preserving the trajectory-level reward ordering.
The expectation emphasizes that dependency estimation, reward normalization, and token gating are recomputed for every rollout batch. The denominator remains the full completion length $T_i$, matching the original GRPO normalization; unselected tokens therefore reduce the effective policy-gradient mass instead of changing the reduction rule. Consequently, \modelname changes the credit resolution of policy optimization: sequence-level summarization rewards are preserved, but their learning pressure is routed toward source-dependent trajectories and tokens.

\begin{table*}[th]
  \centering
  \caption{The results of multi-dimensional evaluation on the CNN/DailyMail dataset. Top three results for each metric are highlighted as \colorbox{bestLight}{\textbf{best}}, \colorbox{secondLight}{second}, and \colorbox{thirdLight}{third}, respectively.}
  \label{tab:combined_results}
  \tiny
  \resizebox{.9\textwidth}{!}{%
  \begin{tabular}{c l c c c c c c c}
    \toprule
    \textbf{Dataset}
    & \textbf{Model}
    & \textbf{Method}
    & \textbf{Coherence} $\uparrow$
    & \textbf{Consistency} $\uparrow$
    & \textbf{Fluency} $\uparrow$
    & \textbf{Relevance} $\uparrow$
    & \textbf{Overall} $\uparrow$
    & \textbf{STD} $\downarrow$ \\
    \midrule

    \multirow{18}{*}{\rotatebox{90}{\textbf{CNN/DailyMail~\cite{nallapati2016abstractive}}}}
    & PEGASUS
    & SFT
    & 0.936
    & 0.939
    & 0.815
    & 0.684
    & 0.843
    & 0.121 \\

    \cmidrule(lr){2-9}
    & \multicolumn{8}{c}{\cellcolor{gray!10}\textit{Scaling Models}} \\
    \cmidrule(lr){2-9}

    & Qwen2.5 1.5B
    & Zero-shot
    & 0.871
    & 0.819
    & 0.936
    & 0.861
    & 0.872
    & 0.048 \\

    & Qwen2.5 7B
    & Zero-shot
    & 0.890
    & 0.820
    & 0.932
    & 0.874
    & 0.879
    & 0.046 \\

    & Qwen2.5 14B
    & Zero-shot
    & 0.931
    & 0.826
    & 0.859
    & 0.907
    & 0.881
    & 0.047 \\

    & Qwen2.5 32B
    & Zero-shot
    & 0.918
    & 0.843
    & 0.933
    & 0.893
    & 0.897
    & 0.040 \\

    \cmidrule(lr){2-9}
    & \multicolumn{8}{c}{\cellcolor{gray!10}\textit{Proprietary Models}} \\
    \cmidrule(lr){2-9}

    & GPT-4
    & Zero-shot
    & \cellcolor{bestLight}\textbf{0.967}
    & 0.840
    & 0.945
    & 0.934
    & 0.921
    & 0.056 \\

    & GPT-5-nano
    & Zero-shot
    & 0.835
    & 0.735
    & 0.813
    & 0.794
    & 0.794
    & 0.084 \\

    \cmidrule(lr){2-9}
    & \multicolumn{8}{c}{\cellcolor{gray!10}\textit{Reinforcement Learning Methods}} \\
    \cmidrule(lr){2-9}

    & Qwen2.5 7B
    & GRPO~\cite{shao2024deepseekmath}
    & 0.908
    & 0.903
    & 0.922
    & 0.954
    & 0.922
    & \cellcolor{thirdLight}0.023 \\

    & Qwen2.5 7B
    & HVO~\cite{song2026balancing}
    & \cellcolor{secondLight}0.961
    & 0.926
    & 0.951
    & 0.934
    & 0.943
    & \cellcolor{bestLight}\textbf{0.016} \\

    & Qwen2.5 7B
    & DAPO~\cite{yu2026dapo}
    & 0.937
    & 0.864
    & 0.906
    & 0.929
    & 0.909
    & 0.036 \\

    & Qwen2.5 7B
    & SAPO~\cite{gao2025soft}
    & 0.931
    & 0.869
    & 0.914
    & 0.927
    & 0.910
    & 0.037 \\

    \cmidrule(lr){2-9}
    & \multicolumn{8}{c}{\cellcolor{gray!10}\textit{Our Method and Ablation Studies}} \\
    \cmidrule(lr){2-9}

    & Qwen2.5 7B
    & \modelname (Ours)-mask 20\%
    & \cellcolor{thirdLight}0.943
    & \cellcolor{secondLight}0.943
    & \cellcolor{bestLight}\textbf{0.980}
    & \cellcolor{bestLight}\textbf{0.969}
    & \cellcolor{secondLight}0.959
    & 0.059 \\

    & Qwen2.5 7B
    & \modelname (Ours)-mask 80\%
    & 0.942
    & \cellcolor{thirdLight}0.940
    & \cellcolor{secondLight}0.979
    & \cellcolor{secondLight}0.968
    & \cellcolor{thirdLight}0.957
    & 0.063 \\

    & Qwen2.5 7B
    & \textbf{\modelname (Ours)}
    & \cellcolor{secondLight}0.961
    & \cellcolor{bestLight}\textbf{0.946}
    & \cellcolor{thirdLight}0.967
    & \cellcolor{thirdLight}0.966
    & \cellcolor{bestLight}\textbf{0.960}
    & \cellcolor{secondLight}0.020 \\

    \bottomrule
  \end{tabular}%
  }
  \vspace{-10pt}
\end{table*}

\section{Experiments}\label{sec:experiments}
\subsection{Experimental Settings}
\noindent\textbf{Datasets}. 
Following~\cite{song2026balancing}, we evaluate \modelname on the CNN/DailyMail abstractive summarization benchmark\footnote{\scriptsize\url{https://huggingface.co/google/pegasus-cnn_dailymail}}~\cite{nallapati2016abstractive}. CNN/DailyMail contains online news articles paired with ordered multi-sentence highlights, and is widely used to test whether a system can compress long news documents while preserving salient events and entities.

\noindent\textbf{Baselines and Evaluation Metrics}.
We compare \modelname with three groups of representative baselines. The first group contains PEGASUS~\cite{zhang2020pegasus}, a supervised abstractive summarization model. Following HVO~\cite{song2026balancing}, we report its SFT result and the zero-shot result of GPT-4~\cite{achiam2023gpt}. The second group includes zero-shot open-source LLMs at different scales, including Qwen2.5 1.5B, 7B, 14B, and 32B~\cite{bai2025qwen2}, which allows us to separate the effect of model scale from that of policy optimization. We further evaluate GPT-5-nano through the Poe API\footnote{\scriptsize\url{https://poe.com/api}}. The third group includes RL-based methods, including GRPO~\cite{shao2024deepseekmath}, HVO~\cite{song2026balancing}, DAPO~\cite{yu2026dapo}, and SAPO~\cite{gao2025soft}. For evaluation, we use UniEval~\cite{zhong2022towards}. We report four standard summarization dimensions: coherence, consistency, fluency, and relevance. The overall score is the arithmetic mean of the four dimensions. We also report STD, the standard deviation over the four dimension scores, to measure whether a method improves summary quality in a balanced manner. Higher values indicate better performance for all UniEval scores, while lower values are better for STD.

\noindent\textbf{Implementation Details}. 
All RL methods use Qwen2.5-7B-Instruct as the initial policy for a fair comparison. We use the same prompt template for training and inference: ``Summarize the Text without any Explanation.'' The maximum prompt length and completion length are set to 2048 and 512 tokens, respectively. We use UniEval-sum to compute the four reward dimensions, and normalize rewards within each generation group. The group size is set to 8. We train for 4 epochs with AdamW~\cite{loshchilov2019decoupled}, using a learning rate of $5\times10^{-7}$, betas of $(0.9,0.999)$, weight decay of $0.1$, a cosine learning-rate schedule, a warmup ratio of $0.1$, and a maximum gradient norm of $0.4$. The rollout temperature is set to 1.0. For \modelname, we randomly mask 50\% of source tokens and compare the token probabilities of the same generated summary under the original and masked sources to estimate source-dependent token importance. The token-importance signal is used only for credit assignment. We reweight each trajectory advantage by its mean token importance normalized by the rollout mean, and apply policy-gradient updates only to the top 40\% most important valid output tokens in each trajectory. At inference time, all trained and zero-shot open-source models use greedy decoding with a maximum of 512 new tokens. Experiments are conducted on 32 $\times $ 4 NVIDIA A100 GPUs with 40GB memory.

\subsection{Performance Comparison}
Table~\ref{tab:combined_results} reports the main evaluation results. \textit{1)} Overall, \modelname achieves the best overall score of 0.960 and ranks first on consistency, fluency, and relevance, reaching 0.946, 0.967, and 0.966, respectively. It also obtains the second-best coherence score of 0.961 and the second-lowest STD of 0.020. Compared with the Qwen2.5 7B zero-shot, \modelname improves the overall score by 0.081 and consistency by 0.126, showing that the gain is not merely inherited from the foundation model but comes from policy optimization. Compared with the larger Qwen2.5 32B zero-shot model, the 7B policy optimized by \modelname still improves the overall score by 0.063, indicating that source-aware RL can be more effective than simply increasing model scale under the same evaluation setting. \textit{2)} Scaling open-source LLMs improves summarization quality only moderately. The overall score increases from 0.872 for Qwen2.5 1.5B to 0.897 for Qwen2.5 32B, but the improvements are uneven across dimensions. In contrast, \modelname substantially improves all dimensions over the Qwen2.5 7B zero-shot model. Against proprietary models, GPT-4 obtains the highest coherence score of 0.967, but \modelname surpasses it on consistency, fluency, relevance, and overall score by 0.106, 0.022, 0.032, and 0.039, respectively. These results suggest that a 7B model can reach highly competitive summarization quality when the optimization signal is assigned to source-dependent tokens. \textit{3)} Among RL-based baselines, \modelname also shows clear advantages. It improves the overall score over GRPO, HVO, DAPO, and SAPO by 0.038, 0.017, 0.051, and 0.050. Compared with HVO, the strongest previous RL baseline in overall performance, \modelname matches its coherence score and further improves consistency, fluency, and relevance by 0.020, 0.016, and 0.032. Although HVO obtains a slightly lower STD, \modelname maintains a similarly balanced profile while achieving much higher absolute quality. This supports our central hypothesis: under sequence-level summarization rewards, explicitly resolving trajectory-token credit through source dependency leads to more reliable improvements than broadcasting the same advantage to all generated tokens.

\subsection{Ablation Study}
Table~\ref{tab:combined_results} also evaluates the masking ratio used in source-dependency estimation. Both 20\% and 80\% masking remain strong, obtaining overall scores of 0.959 and 0.957, respectively, which confirms that the token-importance mechanism is robust to perturbation strength. However, the default 50\% setting achieves the best overall score (0.960), the highest consistency (0.946), and a much lower STD. This indicates that moderate perturbation provides more balanced source-dependency signals than too weak or too aggressive masking.

\subsection{Case Study}
\begin{figure}[t!]
  \begin{center}
\includegraphics[width=.8\linewidth]{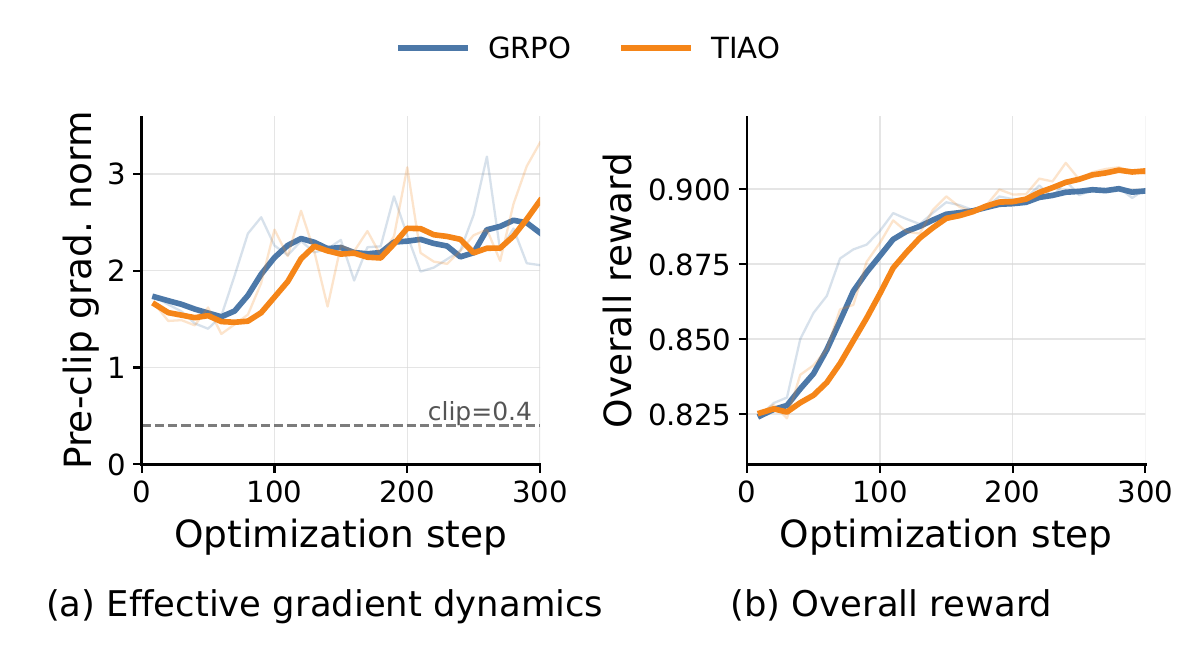}
  \end{center}
    \vspace{-20pt}
  \caption{Case study on effective gradient dynamics and overall reward.}
  \vspace{-10pt}
  \label{fig:case-study}
  \vspace{-7pt}
\end{figure}
Fig.~\ref{fig:case-study} illustrates why \modelname behaves differently from vanilla GRPO. In GRPO, the group-relative advantage is broadcast to all valid tokens in a completion. 
\modelname first estimates how much each sampled token depends on the source document by comparing token probabilities under the original and masked sources, and then uses this dependency signal to reshape trajectory advantages and select the most source-sensitive output tokens for policy-gradient updates. 
GRPO improves the overall reward rapidly in the early stage.
As training proceeds, \modelname shows a stronger pre-clip gradient signal and reaches a higher overall reward. This late-stage separation suggests that concentrating optimization on source-dependent tokens helps the policy keep improving evidence-related decisions instead of repeatedly reinforcing low-information positions. 

\section{Conclusion}\label{sec:conclusion}
In this paper, we address the credit-resolution mismatch in RL-based text summarization, where sequence-level rewards are broadcast to undifferentiated tokens. We propose \modelname, a token importance-aware policy optimization method that estimates source-dependent token importance through source masking, reweights trajectory advantages, and focuses updates on important output tokens. Experiments on CNN/DailyMail show that the 7B policy achieves competitive quality against strong open-source and proprietary baselines.

\clearpage
\bibliographystyle{IEEEbib}
\bibliography{strings}
\end{document}